\documentclass[10pt,twocolumn,letterpaper]{article}

\usepackage{cvpr}
\usepackage{times}
\usepackage{epsfig}
\usepackage{graphicx}
\usepackage{rotating}
\usepackage{multirow}
\usepackage[table]{xcolor}
\usepackage{amsmath}
\usepackage[labelfont={bf,footnotesize},font=footnotesize]{caption}
\usepackage{amssymb}
\usepackage{pifont}
\usepackage[pagebackref=true,breaklinks=true,letterpaper=true,colorlinks,bookmarks=false,citecolor=blue]{hyperref}

\cvprfinalcopy 

\def\cvprPaperID{3856} 

\ifcvprfinal\fi

\begin{document}

\title{AGA : Attribute-Guided Augmentation}

\author{Mandar Dixit\\
UC San Diego\\
{\tt\small mdixit@ucsd.edu}
\and
Roland Kwitt\\
University of Salzburg\\
{\tt\small rkwitt@gmx.at}
\and
Marc Niethammer\\
UNC Chapel Hill\\
{\tt\small mn@cs.unc.edu}
\and
Nuno Vasconcelos\\
UC San Diego\\
{\tt\small nvasconcelos@ucsd.edu}
}

\maketitle

\begin{abstract}
We consider the problem of data augmentation, i.e., generating artificial
samples to extend a given corpus of training data. Specifically, we propose
attributed-guided augmentation (AGA) which learns a mapping
that allows synthesis of data 
 such that an attribute of a synthesized sample 
is at a desired value or strength. This is particularly interesting in situations 
where little data with no attribute annotation is available for learning, but 
we have access to an 
external corpus of heavily annotated samples. 
While prior works primarily augment in the space of images,
we propose to perform augmentation in feature space instead.
We implement our approach as a deep encoder-decoder architecture that 
learns the synthesis function in an end-to-end manner. 
We demonstrate the utility of our approach on the problems of (1) 
one-shot object recognition in a transfer-learning setting where
we have no prior knowledge of the new classes, as well as 
(2) object-based one-shot scene recognition.
As external data, we leverage 3D depth and pose information from the 
SUN RGB-D dataset. Our experiments show that attribute-guided augmentation of
high-level CNN features considerably improves one-shot recognition 
performance on both problems.
\end{abstract}

\section{Introduction}
\label{section:introduction}

Convolutional neural networks~(CNNs), trained on large scale data, have significantly advanced the state-of-the-art 
on traditional vision problems such as object recognition~\cite{Krizhevsky12a,Simonyan14a,Szegedy15a} and 
object detection~\cite{Girshick15a,Ren15a}. Success of these networks is mainly due to their high selectivity 
for semantically meaningful visual concepts, \eg, objects and object parts~\cite{Fergus14a}.
In addition to ensuring good performance on the problem of interest, this property of CNNs also allows 
for {\it transfer\/} of knowledge to several other vision tasks~\cite{Donahue14a,Gong14a,Cimpoi15a,Dixit15a}. 
The object recognition network of~\cite{Krizhevsky12a}, \eg, has been successfully used for object 
detection~\cite{Girshick15a,Ren15a}, scene classification~\cite{Gong14a,Dixit15a}, texture 
classification~\cite{Cimpoi15a} and domain adaptation~\cite{Donahue14a}, using various 
transfer mechanisms. 

\begin{figure}[t!]
\centering{
\includegraphics[width=0.99\columnwidth]{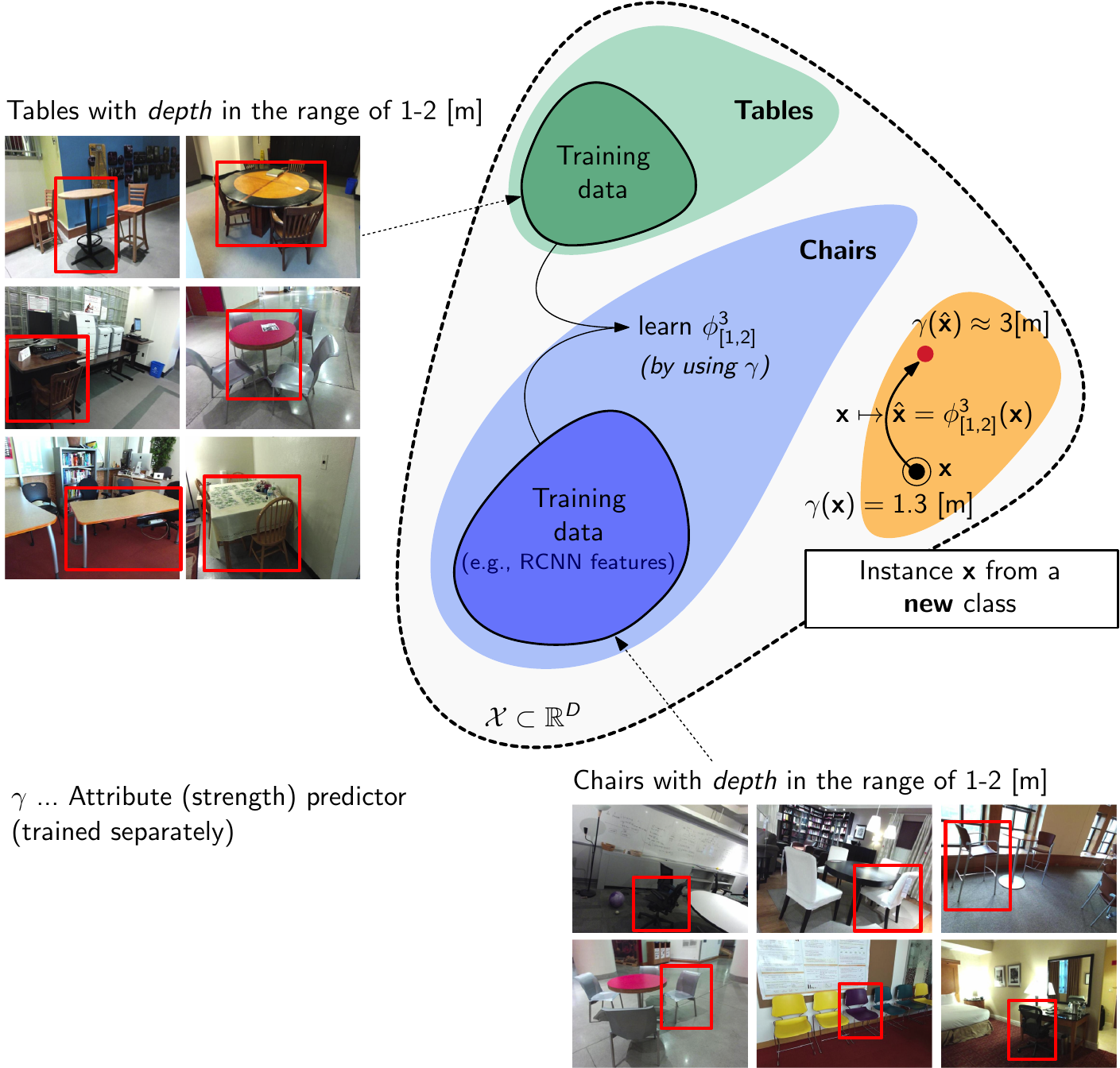}
}
\caption{\label{fig:intro} Given a predictor $\gamma: \mathcal{X} \to \mathbb{R}_+$
of some object attribute (\eg, depth or pose), we propose to \emph{learn} a mapping 
of object features $\mathbf{x} \in \mathcal{X}$, such that
(1) the new synthetic feature $\hat{\mathbf{x}}$ is ``close'' 
to $\mathbf{x}$ (to preserve object identity) and (2) the predicted 
attribute value $\gamma(\hat{\mathbf{x}}) = \hat{t}$ of $\hat{\mathbf{x}}$
matches a desired object attribute value $t$, \ie, $t-\hat{t}$ is small. In this illustration,
we learn a mapping for features with associated \emph{depth} values in the
range of 1-2 [m] to $t=3$~[m] and apply this mapping to an instance of a new 
object class. In our approach, this mapping is learned in 
an \emph{object-agnostic} manner. With respect to our example, this means that
\emph{all} training data from `chairs' and `tables' is used to a learn feature 
synthesis function $\phi$.}
\end{figure}

CNN-based transfer is generally achieved either by {\it finetuning\/} a pre-trained network,  
such as in~\cite{Krizhevsky12a}, on a new image dataset or by designing a new image 
representation on such a dataset based on the activations of the pre-trained network 
layers~\cite{Donahue14a,Gong14a,Dixit15a,Cimpoi15a}.
Recent proposals of transfer have shown highly competitive performance on different predictive 
tasks with a modest amount of new data (as few as 50 images per class). The effectiveness 
of transfer-based methods, however, has not yet been tested under more severe constraints such as 
in a {\it few-shot\/} or a {\it one-shot\/} learning scenario. In these problems, the number 
of examples available for learning may be as few as one per class. Fine-tuning a pre-trained 
CNN with millions of parameters to such inadequate datasets is clearly not a viable option. 
A one-shot classifier trained on CNN activations will also be prone to over-fitting due to the 
high dimensionality of the feature space. The only way to solve the problem of limited data is 
to {\it augment\/} the training corpus by obtaining more examples for the given classes.

While augmentation techniques can be as simple as flipping, rotating, adding noise, or 
extracting random crops from images \cite{Krizhevsky12a, Chatfield14, Zeiler14a}, 
\emph{task-specific}, or \emph{guided} augmentation strategies \cite{Charalambous16a,
Hauberg16a,Rogez16a,Peng15a} have the potential to generate more realistic synthetic 
samples. This is a particularly important issue, since performance of CNNs heavily
relies on sufficient coverage of the variability that we expect in 
unseen testing data. In scene recognition, we desire, for example, sufficient variability in the 
constellation and transient states of scene categories (\cf \cite{Kwitt16a}), whereas in object recognition, 
we desire variability in the specific incarnations of certain objects, lighting 
conditions, pose, or depth, just to name a few. Unfortunately, this variability 
is often dataset-specific and can cause substantial bias in recognition results 
\cite{Torralba11a}. 

An important observation in the context of our work is 
that augmentation is typically performed on the image, or video level.
While this is not a problem with simple techniques, such as flipping or cropping, it can 
become computationally expensive if more elaborate augmentation 
techniques are used. We argue that, in specific problem settings, 
augmentation might as well be performed in \emph{feature space}, 
especially in situations where features are input to subsequent 
learning steps. This is common, \eg, in recognition tasks, where
the softmax output of trained CNNs is often not used directly, but
activations at earlier layers are input to an 
external discriminant classifier. 

 




\vskip0.5ex
\noindent
\textbf{Contribution.} 
We propose an approach to augment the training set with \emph{feature descriptors}
instead of images. Specifically, we advocate an augmentation technique
that learns to synthesize features, guided by desired values
for a set of object attributes, such as depth or pose. 
An illustration of this concept is shown in Fig.~\ref{fig:intro}.
We first train a fast RCNN~\cite{Girshick15a} 
detector to identify objects in 2D images. This is followed by training a 
neural network regressor which predicts the 3D attributes of a detected object, 
namely its depth from the camera plane and pose. 
An encoder-decoder network is then trained which, for a detected object at a certain 
depth and pose, will ``hallucinate'' the changes in its RCNN features for 
a set of desired depths/poses. Using this architecture, for a new image, we are able to 
augment existing feature descriptors by 
an auxiliary set of features that correspond to the object changing its 
3D position. Since our framework relies on object attributes to 
guide augmentation, we refer to it as 
{\it attribute-guided augmentation (AGA)\/}.

%
%

\vskip0.5ex
\noindent
\textbf{Organization.} Sec.~\ref{section:relatedwork}\ reviews
prior work. Sec.~\ref{section:architecture} introduces
the proposed encoder-decoder architecture for attribute-guided 
augmentation. Sec.~\ref{section:experiments} studies 
the building blocks of this approach in detail and  
demonstrates that AGA in feature space improves 
one-shot object recognition and object-based scene 
recognition performance on previously unseen classes. Sec.~\ref{section:discussion} concludes the paper with
a discussion and an outlook on potential future directions.

\section{Related work}
\label{section:relatedwork}

Our review of related work primarily focuses on 
\emph{data augmentation} strategies. While many techniques
have been proposed in the context of training deep 
neural networks to avoid over-fitting and to increase variability
in the data, other (sometimes closely related) 
techniques have previously appeared in the context
of one-shot and transfer learning. We can roughly
group existing techniques into (1) \emph{generic}, 
computationally cheap approaches and (2) task-specific, 
or guided approaches that are typically more 
computationally involved.

As a representative of the first group, Krizhevsky \etal \cite{Krizhevsky12a} 
leverage a set of label-preserving transformations, such 
as patch extraction + reflections, and PCA-based intensity
transformations, to increase training sample size. Similar techniques
are used by Zeiler and Fergus \cite{Zeiler14a}. In \cite{Chatfield14},
Chatfield and Zisserman demonstrate that
the augmentation techniques of \cite{Krizhevsky12a}
are not only beneficial for training deep architectures, but 
shallow learning approaches equally benefit from
such \emph{simple} and \emph{generic} schemes.

In the second category of guided-augmentation techniques,
many approaches have recently been proposed.
In \cite{Charalambous16a}, \eg, Charalambous and Bharath
employ guided-augmentation in the context of
gait recognition. The authors suggest to simulate synthetic
gait video data (obtained from avatars) with respect to 
various confounding factors (such as clothing, hair, etc.) 
to extend the training corpus. Similar in spirit, Rogez and 
Schmid \cite{Rogez16a} propose an image-based
synthesis engine for augmenting existing 2D human pose
data by photorealistic images with greater pose variability.
This is done by leveraging 3D motion capture (MoCap) data. 
In \cite{Peng15a}, Peng \etal also use 3D data, in 
the form of CAD models, to render synthetic images 
of objects (with varying pose, texture, background) that 
are then used to train CNNs for object detection. It is shown that
synthetic data is beneficial, especially in situations where few
(or no) training instances are available, but 3D CAD models
are. Su \etal \cite{Su15a} follow a similar pipeline
of rendering images from 3D models for viewpoint 
estimation, however, with substantially more synthetic data
obtained, \eg, by deforming existing 3D models
before rendering.

Another (data-driven) guided augmentation technique is 
introduced  by Hauberg \etal \cite{Hauberg16a}. The authors 
propose to \emph{learn} class-specific transformations 
from external training data, instead of manually specifying 
transformations as in \cite{Krizhevsky12a,Zeiler14a,Chatfield14}. 
The learned transformations are then applied to the samples of 
each class. Specifically, diffeomorphisms are learned
from data and encouraging results are demonstrated in the 
context of digit recognition on MNIST. Notably, this 
strategy is conceptually similar to earlier work by 
Miller \etal \cite{Miller00a} on one-shot learning, where 
the authors synthesize additional data for digit images 
via an iterative process, called \emph{congealing}. During 
that process, external images of a given category are aligned by
optimizing over a class of geometric transforms (\eg, 
affine transforms). These transformations are then applied
to single instances of the new classes to increase 
data for one-shot learning.

Marginally related to our work, we remark that alternative 
approaches to implicitly learn spatial transformations have
been proposed. For instance, Jaderberg \etal \cite{Jaderberg15a}  
introduce \emph{spatial transformer} modules that can be
injected into existing deep architectures to implicitly capture 
spatial transformations inherent in the data, thereby improving
invariance to this class of transformations.

While \emph{all} previously discussed methods essentially 
propose \emph{image-level} augmentation to train CNNs, 
our approach is different in that we perform
augmentation in \emph{feature space}. Along these lines, the 
approach of Kwitt \etal \cite{Kwitt16a} is conceptually 
similar to our work. In detail, the authors suggest to learn
how features change as a function of the strength of certain 
transient attributes (such as sunny, cloudy, or foggy) in 
a scene-recognition context. These models are 
then transferred to previously unseen data for one-shot recognition. There are, however, two key differences between their approach and ours. First, they require datasets labeled with \emph{attribute trajectories}, \ie, all variations of an attribute for every instance of a class. We, on the other hand, make use of conventional datasets that seldom carry such extensive labeling. Second, their augmenters are simple linear regressors trained in a \emph{scene-class specific} manner. In contrast, we learn deep non-linear models in a \emph{class-agnostic} manner which enables a straightforward application to recognition in transfer settings. 

\section{Architecture}
\label{section:architecture}

\noindent
\textbf{Notation.}
To describe our architecture, we let $\mathcal{X}$ denote 
our feature space, $\mathbf{x} \in \mathcal{X} \subset \mathbb{R}^D$ denotes a
feature descriptor (\eg, a representation of an object) and 
$\mathcal{A}$ denotes a set of attributes that are available for
objects in the external training corpus. Further, we let $s \in \mathbb{R}_+$ denote 
the value of an attribute $A \in \mathcal{A}$, associated with 
$\mathbf{x}$. We assume (1) that this attribute can be predicted by
an attribute regressor $\gamma: \mathcal{X} \rightarrow \mathbb{R}_+$ 
and (2) that its range can be divided into 
$I$ intervals $[l_i,h_i]$, where $l_i,h_i$ denote the lower 
and upper bounds of the $i$-th interval. The set of desired
object attribute values is $\{t_1,\ldots,t_T\}$.

\vskip0.5ex
\noindent
\textbf{Objective.}
On a conceptual level, we aim for a synthesis function $\phi$ which,
given a desired value $t$ for some object attribute $A$, 
transforms the object features $\mathbf{x} \in \mathcal{X}$ such that the attribute strength changes in a controlled manner 
to $t$. More formally, we aim to learn
\begin{equation}
\phi: \mathcal{X} \times \mathbb{R}_+ \rightarrow 
\mathcal{X},\ (\mathbf{x},t)\mapsto \hat{\mathbf{x}}, \quad \text{s.t.}\quad 
\gamma(\hat{\mathbf{x}}) \approx t\enspace. 
\label{eqn:general}
\end{equation}
Since, the formulation in Eq.~\eqref{eqn:general} 
is overly generic, we constrain the 
problem to the case where we learn different
$\phi_i^k$ for a selection of intervals $[l_i,h_i]$
within the range of attribute $A$ and a selection of 
$T$ desired object attribute values $t_k$. In our
illustration of Fig.~\ref{fig:intro}, \eg, we have 
one interval $[l,h]=[1,2]$ and one attribute (depth)
with target value 3[m]. While
learning separate synthesis functions simplifies 
the problem, it requires a good a-priori \emph{attribute 
(strength) predictor}, since, otherwise, we could not decide which 
$\phi_i^k$ to use. During testing, we (1) predict the 
object's attribute value from its original feature $\mathbf{x}$, 
\ie, $\gamma(\mathbf{x}) =\hat{t}$, and then (2) synthesize additional 
features as $\hat{\mathbf{x}} = 
\phi_i^k(\mathbf{x})$ for $k=1,\ldots,T$. If $\hat{t} \in [l_i,h_i]
\wedge t_k \notin [l_i,h_i]$, $\phi_i^k$ is used. 
Next, we discuss each component of this approach 
in detail.

\subsection{Attribute regression}
\label{subsection:covariateregression}

\begin{figure}[h!]
\centering{
\includegraphics[scale=0.78]{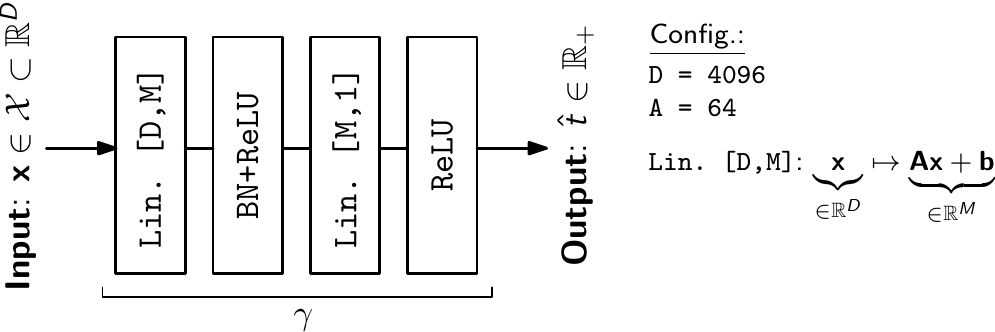}}
\caption{Architecture of the attribute regressor $\gamma$.
\label{fig:COR}}
\end{figure}

An essential part of our architecture is the attribute
regressor $\gamma: \mathcal{X} \rightarrow \mathbb{R}_+$ 
for a given attribute $A$. This regressor takes as input a feature 
$\mathbf{x}$ and predicts its strength or value, 
\ie, $\gamma(\mathbf{x}) = \hat{t}$. While $\gamma$ could, 
in principle, be implemented by a variety of approaches, such 
as support vector regression \cite{Drucker97a} or Gaussian processes \cite{Rasmussen05a}, 
we use a two-layer neural network instead, to accomplish this task. This is not an arbitrary 
choice, as it will later enable us to easily re-use this building 
block in the learning stage of the synthesis function(s) $\phi_i^k$.
The architecture of the attribute regressor is shown in 
Fig.~\ref{fig:COR}, consisting of two linear layers, 
interleaved by batch normalization (BN) \cite{Ioffe15a} and rectified linear 
units (ReLU) \cite{Nair10a}. While this architecture is 
admittedly simple, adding more layers did not lead to 
significantly better results in our experiments. Nevertheless,
the design of this component is problem-specific and could
easily be replaced by more complex variants, depending on
the characteristics of the attributes that need to be predicted.

\vskip0.5ex
\noindent
\textbf{Learning.} The attribute regressor can easily
be trained from a collection of $N$ training tuples $
\{(\mathbf{x}_i,s_i)\}_{i=1}^N$ for each attribute.
As the task of the attribute regressor is to predict 
in which interval the original feature $\mathbf{x}$ 
resides, we do not need to organize the training data 
into intervals in this step.

\subsection{Feature regression}
\label{subsection:aug}

To implement\footnote{We omit the sub-/superscripts for readability.} $\phi$, we design an encoder-decoder
architecture, reminiscent of a conventional autoencoder \cite{Bengio09a}.
Our objective, however, is not to encode and then reconstruct 
the input, but to produce an output that 
resembles a feature descriptor of an object at a 
desired attribute value.

\begin{figure}[t!]
\centering{
\includegraphics[width=\columnwidth]{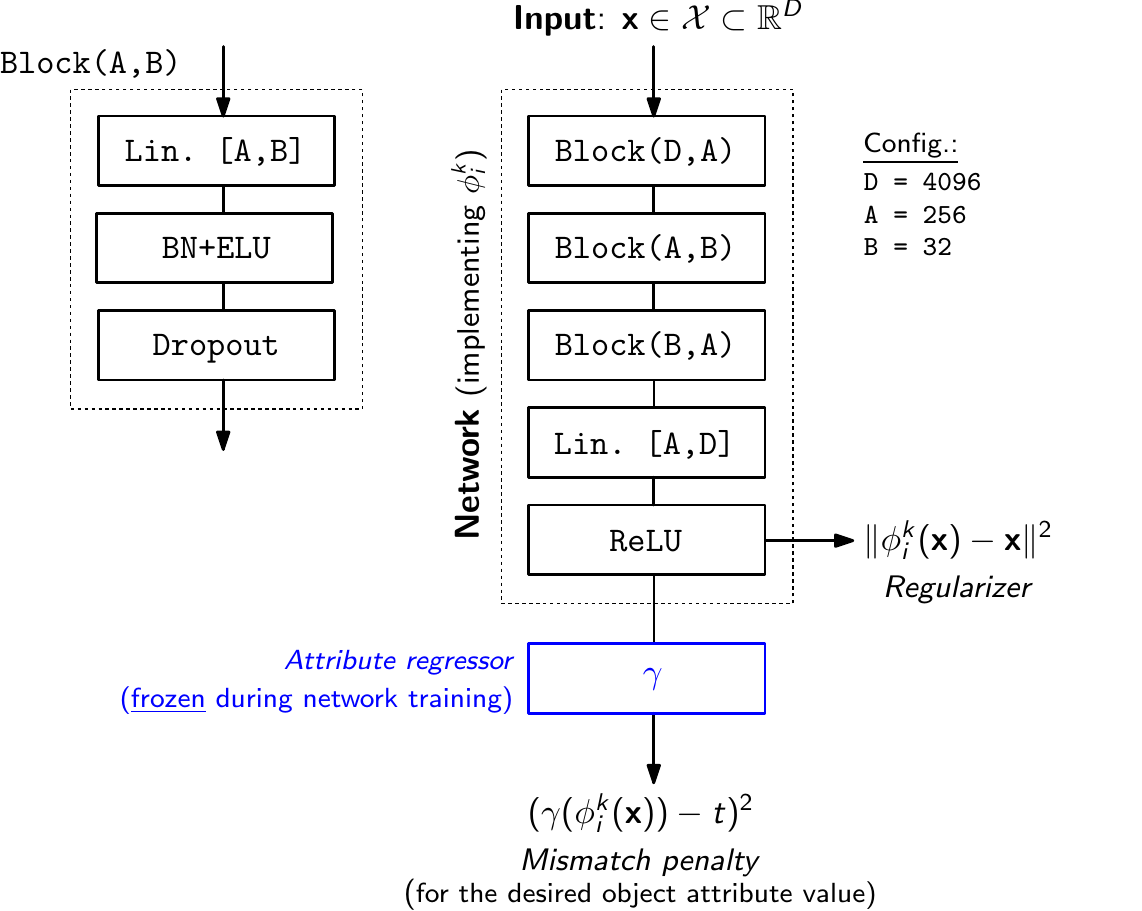}}
\caption{Illustration of the proposed encoder-decoder network for AGA.
During \emph{training}, the attribute regressor $\gamma$ is appended to
the network, whereas, for \emph{testing} (\ie, feature synthesis) 
this part is removed. When learning $\phi_i^k$, the input $\mathbf{x}$ 
is such that the associated attribute value $s$ is within $[l_i,h_i]$ 
and one $\phi_i^k$ is learned per desired attribute value $t_k$.
\label{fig:EDN}}
\end{figure}

In other words, the \emph{encoder} essentially learns to extract
the essence of features; the \emph{decoder}
then takes the encoding and decodes it to the desired result. In
general, we can formulate the optimization problem as
\begin{equation}
\min_{\phi \in \mathcal{C}} L(\mathbf{x},t; \phi) = (\gamma(\phi(\mathbf{x}))-t)^2\enspace,
\label{eqn:unconstrained}
\end{equation}
where the minimization is 
over a suitable class of functions $\mathcal{C}$. Notably, when 
implementing $\phi$ as an encoder-decoder network with an 
appended (pre-trained) attribute predictor (see Fig.~\ref{fig:EDN})
and loss $(\gamma(\phi(\mathbf{x}))-t)^2$, we have little control 
over the decoding result in the sense that we cannot guarantee 
that the \emph{identity} of the input is preserved. This means
that features from a particular object class might map to 
features that are no longer recognizable as
this class, as the encoder-decoder will \emph{only} learn to ``fool'' the 
attribute predictor $\gamma$.
For that reason, we add a \emph{regularizer} to the objective
of Eq.~\eqref{eqn:unconstrained}, \ie, we require the 
decoding result to be close, \eg, in the 2-norm, to the
input. This changes the optimization problem of Eq.~\eqref{eqn:unconstrained} to
\begin{equation}
\min_{\phi \in \mathcal{C}} L(\mathbf{x},t; \phi) = \underbrace{(\gamma(\phi(\mathbf{x}))-t)^2}_\text{Mismatch penalty} + 
\lambda \underbrace{\| \phi(\mathbf{x}) - \mathbf{x} \|^2}_{\text{Regularizer}}
\enspace.
\label{eqn:constrained}
\end{equation}
Interpreted differently, this resembles the loss of an autoencoder 
network with an added \emph{target attribute mismatch} penalty. 
The encoder-decoder network that implements the function class $\mathcal{C}$ 
to learn $\phi$ is shown in Fig.~\ref{fig:EDN}. 
The core building block is a combination of a linear layer, 
batch normalization, ELU \cite{Clevert16a}, followed by dropout 
\cite{Srivastava14a}. After the final linear
layer, we add one ReLU layer to enforce $\hat{\mathbf{x}} \in \mathbb{R}^D_+$.

\vskip0.5ex
\noindent
\textbf{Learning.} Training the encoder-decoder network of Fig.~\ref{fig:EDN}
requires an a-priori trained attribute regressor $\gamma$ for each given 
attribute $A \in \mathcal{A}$. During training, this
attribute regressor is appended to the network and its 
\emph{weights are frozen}. Hence, only the encoder-decoder weights
are updated. To train one $\phi_i^k$ for each interval $[l_i,h_i]$ 
of the object attribute range and a desired object attribute value $t_k$, we partition
the training data from the external corpus into subsets $\mathcal{S}_i$, 
such that $\forall (\mathbf{x}_n,s_n) \in \mathcal{S}_i: s_n \in [l_i,h_i]$. 
One $\phi_i^k$ is learned from $\mathcal{S}_i$ for each desired
object attribute value $t_k$. As 
training is in feature space $\mathcal{X}$, we 
have no convolutional layers and consequently training is 
computationally cheap. For testing, the attribute regressor 
is removed and only the trained encoder-decoder network (implementing 
$\phi_i^k$) is used to synthesize features. Consequently, given 
$|\mathcal{A}|$ attributes, $I$ intervals per attribute and 
$T$ target values for an object attribute, we obtain 
$|\mathcal{A}|\cdot I\cdot T$ synthesis functions.

\section{Experiments}
\label{section:experiments}

We first discuss the generation of adequate training data for the 
encoder-decoder network, then evaluate every component of our architecture 
separately and eventually demonstrate its utility on (1) one-shot object recognition 
in a transfer learning setting and (2) one-shot scene recognition.

\vskip0.5ex
\noindent
\textbf{Dataset.} We use the SUN {RGB-D} dataset from 
Song \etal \cite{Song15a}. This dataset contains 10335 
RGB images with depth maps, as well as detailed 
annotations for more than 1000 objects in the form of 
2D and 3D bounding boxes.
In our setup, we use object depth and pose as our
attributes, \ie, $\mathcal{A} = \{\texttt{Depth}, 
\texttt{Pose}\}$. For each ground-truth 3D bounding box, 
we extract the depth value at its centroid and obtain 
pose information as the rotation of the 3D bounding
box about the vertical $y$-axis. In all experiments, we use the first
5335 images as our \emph{external database}, \ie, the database
for which we assume availability of attribute annotations.
The remaining 5000 images are used for testing; more details
are given in the specific experiments.

\begin{figure}
\centering{
\includegraphics[width=\columnwidth]{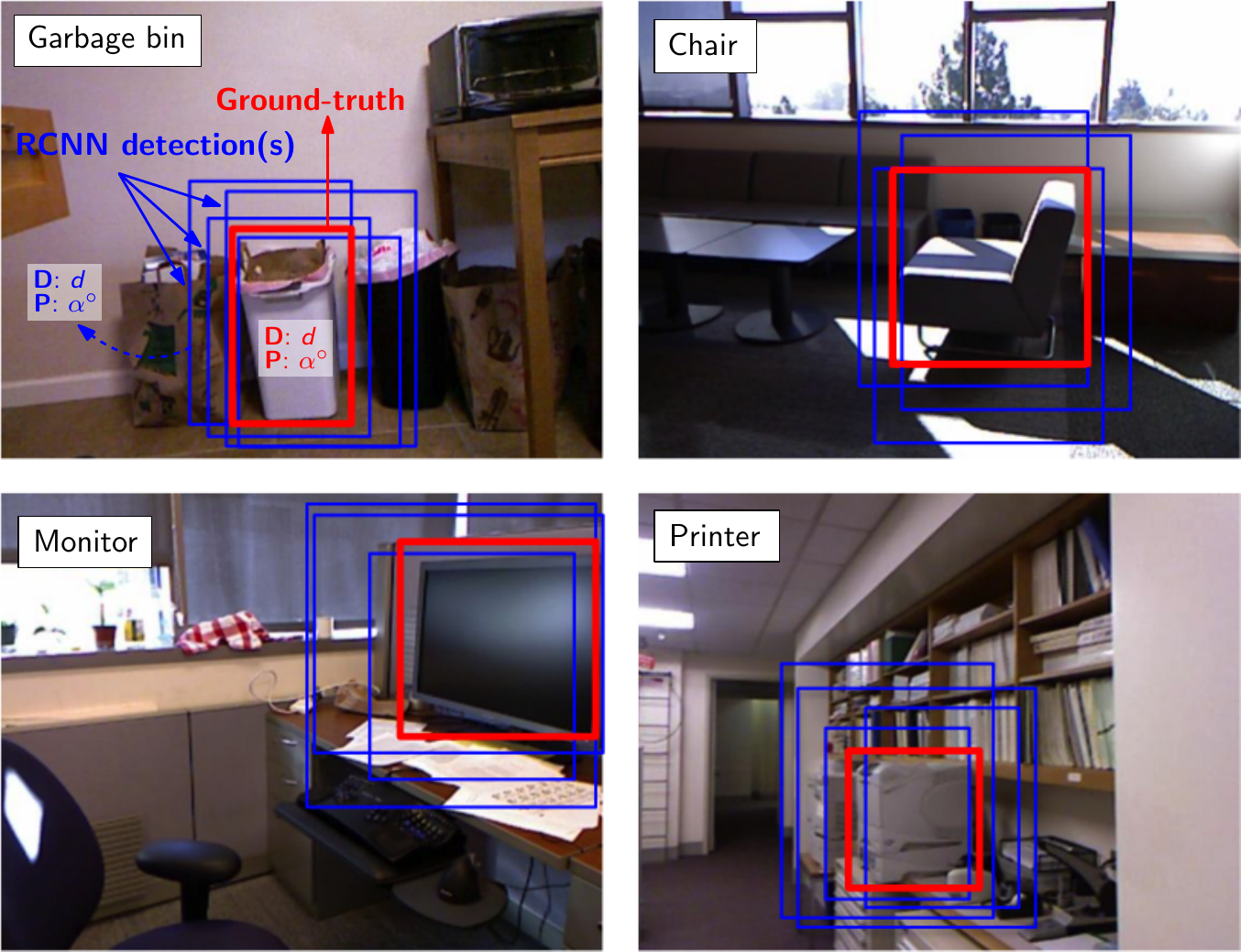}}
\caption{Illustration of \emph{training data generation}. \emph{First}, 
we obtain fast RCNN \cite{Girshick15a} activations (\texttt{FC7} layer)
of Selective Search \cite{Uijlings13a} proposals  
that overlap with 2D ground-truth bounding boxes (IoU $>$ 0.5) 
and scores $>$ 0.7 (for a particular object class) 
to generate a sufficient amount of training data. 
\emph{Second}, attribute values (\ie, depth \textbf{D} and pose 
\textbf{P}) 
of the corresponding 3D ground-truth bounding boxes are 
associated with the proposals (best-viewed in color). \label{fig:trainingdata}}
\end{figure}

\vskip0.5ex
\noindent
\textbf{Training data.} Notably, in SUN RGB-D, the
number of instances of each object class are not 
evenly distributed, simply because this dataset was
not specifically designed for object recognition tasks. 
Consequently, images are also not object-centric, meaning 
that there is substantial variation in the location of 
objects, as well as the depth and pose at which they occur. 
This makes it difficult to extract a sufficient 
and balanced number of feature descriptors per object class, 
if we would \emph{only} use the ground-truth bounding boxes to
extract training data. 
We circumvent this problem by leveraging the fast RCNN detector 
of \cite{Girshick15a} with object proposals generated by Selective
Search \cite{Uijlings13a}. In detail, we finetune the ImageNet 
model from \cite{Girshick15a} to {SUN RGB-D}, using
the same 19 objects as in \cite{Song15a}. We then run 
the detector on all images from our training split and 
keep the proposals with detection scores $>0.7$
and a sufficient overlap (measured by the IoU $>$0.5) 
with the 2D ground-truth bounding boxes. This is a 
simple augmentation technique to increase the amount
of available training data. 
The associated RCNN activations (at the \texttt{FC7} layer) 
are then used as our features $\mathbf{x}$. Each 
proposal that remains after overlap and 
score thresholding is annotated by the attribute information 
of the corresponding ground-truth bounding box in 3D. 
As this strategy generates a larger number of descriptors 
(compared to the number of ground-truth bounding
boxes), we can evenly balance the training data in the sense
that we can select an equal number of detections per object
class to train (1) the attribute regressor and (2) the
encoder-decoder network. Training data generation is illustrated
in Fig.~\ref{fig:trainingdata} on four example images.

\vskip0.5ex
\noindent
\textbf{Implementation.} The attribute regressor and 
the encoder-decoder network are implemented in \texttt{Torch}.
All models are trained using \texttt{Adam} \cite{Kingma15a}. 
For the attribute regressor, we train for 30 epochs with a
batch size of 300 and a learning rate of 
$0.001$. The encoder-decoder network is also trained for 30
epochs with the same learning rate, but with a batch
size of 128. The dropout probability during training is set to 
$0.25$. No dropout is used for testing. For our classification
experiments, we use a linear C-SVM, as implemented in
\texttt{liblinear} \cite{Fan08a}.
On a Linux system, running
Ubuntu 16.04, with 128 GB of memory and one NVIDIA 
Titan X, training one model (\ie, one $\phi_i^k$) takes 
$\approx 30$ seconds. The relatively low demand on 
computational resources highlights the advantage of 
AGA in feature space, as no convolutional layers
need to be trained. All trained models+source code are publicly 
available online\footnote{\url{https://github.com/rkwitt/GuidedAugmentation}}.

\subsection{Attribute regression}
\label{subsection:EvalCovariateRegression}

While our strategy, AGA, to data augmentation is 
\emph{agnostic} to the object classes, in both the
training and testing dataset, it is interesting to 
compare attribute prediction 
performance to the case where we train \emph{object-specific}
regressors. In other words, we compare object-agnostic
training to training one regressor
$\gamma_j,\ j \in \{1,\ldots, |\mathcal{S}|\}$ 
for each object class in $\mathcal{S}$. 
This allows us to quantify the 
potential loss in prediction performance in the 
object-agnostic setting. 

\begin{table}[t!]
\scriptsize
\centering{
\begin{tabular}{rcccc}
\hline
\multirow{ 2}{*}{\textbf{Object}} & \multicolumn{2}{c}{\textbf{D} (MAE [m])}  & \multicolumn{2}{c}{\textbf{P} (MAE [deg])}\\
& per-object & agnostic  & per-object & agnostic \\
\hline
\texttt{             bathtub} & 0.23 & 0.94 & 37.97 & 46.85\\ 
\texttt{                 bed} & 0.39 & 0.30 & 44.36 & 42.59\\ 
\texttt{           bookshelf} & 0.57 & 0.43 & 52.95 & 41.41\\ 
\texttt{                 box} & 0.55 & 0.51 & 27.05 & 38.14\\ 
\texttt{               chair} & 0.37 & 0.31 & 37.90 & 32.86\\ 
\texttt{             counter} & 0.54 & 0.62 & 40.16 & 52.35\\ 
\texttt{                desk} & 0.41 & 0.36 & 48.63 & 41.71\\ 
\texttt{                door} & 0.49 & 1.91 & 52.73 & 102.23\\ 
\texttt{             dresser} & 0.32 & 0.41 & 67.88 & 70.92\\ 
\texttt{         garbage bin} & 0.36 & 0.32 & 47.51 & 45.26\\ 
\texttt{                lamp} & 0.42 & 0.69 & 25.93 & 23.91\\ 
\texttt{             monitor} & 0.24 & 0.22 & 34.04 & 25.85\\ 
\texttt{         night stand} & 0.56 & 0.65 & 23.80 & 20.21\\ 
\texttt{              pillow} & 0.38 & 0.43 & 32.56 & 35.64\\ 
\texttt{                sink} & 0.20 & 0.19 & 56.52 & 45.75\\ 
\texttt{                sofa} & 0.40 & 0.33 & 34.36 & 34.51\\ 
\texttt{               table} & 0.37 & 0.33 & 41.31 & 37.30\\ 
\texttt{                  tv} & 0.35 & 0.48 & 35.29 & 24.23\\ 
\texttt{              toilet} & 0.26 & 0.20 & 25.32 & 19.59\\ 
\hline
$\varnothing$				  &  \cellcolor{black!10}{\textbf{0.39}} &  \cellcolor{black!10}{0.51} & 
 \cellcolor{black!10}{\textbf{40.33}} &  \cellcolor{black!10}{41.12}\\
\hline
              \end{tabular}}
\caption{\label{table:maeCOR} Median-Absolute-Error (MAE), for depth / pose, 
of the attribute regressor, evaluated on \emph{19 objects} from \cite{Song15a}.
In our setup, the pose estimation error quantifies the error in predicting a
rotation around the $z$-axis. \textbf{D} indicates \texttt{Depth}, \textbf{P} indicates
\texttt{Pose}. For reference, the range of of the object attributes in the training data
is [0.2m, 7.5m] for \texttt{Depth} and [0$^\circ$, 180$^\circ$] for \texttt{Pose}. 
Results are averaged over 5 training / evaluation runs.}
\end{table}

Table~\ref{table:maeCOR} lists the median-absolute-error (MAE) of
depth (in [m]) and pose (in [deg]) prediction per object. We train
on instances of 19 object classes ($\mathcal{S}$) in our training 
split of SUN RGB-D and test on instances of the same object classes,
but extracted from the testing split.
As we can see, training in an object-specific manner leads to 
a lower MAE overall, both for depth and pose. 
This is not surprising, since the training data is more specialized 
to each particular object, which essentially amounts to solving 
simpler sub-problems. However, in many cases, 
especially for depth, the object-agnostic regressor performs on par, 
except for object classes with fewer training samples (\eg, door). 
We also remark that, in general, pose estimation from 2D data is
a substantially harder problem than depth estimation (which works
remarkably well, even on a per-pixel level, \cf \cite{Liu15a}). 
Nevertheless, our recognition experiments (in Secs. \ref{subsection:one-shot} and \ref{section:exp_scenes}) show that 
even with mediocre performance of the pose predictor (due to 
symmetry issues, etc.), augmentation along this dimension is still 
beneficial.

\subsection{Feature regression}
\label{subsection:EvalFeatureRegression}

We assess the performance of our regressor(s) 
$\phi_i^k$, shown in Fig.~\ref{fig:EDN}, that are used for synthetic feature generation.
In all experiments, we use an overlapping sliding window to bin
the range of each attribute $A \in \mathcal{A}$ into $I$ 
intervals $[l_i,h_i]$. In case of \texttt{Depth}, we set $[l_0,h_0] = [0,1]$ 
and shift each interval by $0.5$ meter; in case of \texttt{Pose}, we 
set $[l_0,h_0] = [0^\circ,45^\circ]$ and shift by $25^\circ$. 
We generate as many intervals as needed to cover the 
full range of the attribute values in the training data.
The bin-width / step-size were set to ensure a roughly
equal number of features in each bin.
For augmentation, we choose $0.5, 1, \ldots, \max(\texttt{Depth})$ as
target attribute values for \texttt{Depth} and  
$45^\circ, 70^\circ,\ldots, 180^\circ$ for \texttt{Pose}. 
This results in $T=11$ target values for 
\texttt{Depth} and $T=7$ for \texttt{Pose}. 

We use two separate evaluation metrics to assess the performance of
$\phi_i^k$. \emph{First}, we are interested in \emph{how well} the feature
regressor can generate features that correspond to the desired attribute
target values.
To accomplish this, we run each synthetic feature $\hat{\mathbf{x}}$ through
the attribute predictor and assess the MAE, \ie, $|\gamma(\hat{\mathbf{x}}) -t|$, 
over all attribute targets $t$. Table~\ref{table:ednnonseen} lists the average 
MAE, per object, for (1) features from object classes that were \emph{seen} 
in the training data and (2) features from objects that we have never seen 
before. As wee can see from Table \ref{table:ednnonseen}, MAE's for 
seen and unseen objects are similar, indicating that the encoder-decoder 
has learned to synthesize features, such that $\gamma(\hat{\mathbf{x}}) 
\approx t$.

\emph{Second}, we are interested in \emph{how much} synthesized features
\emph{differ} from original features. While we cannot evaluate this directly 
(as we do not have data from one particular object instance at
multiple depths and poses), we can assess how ``close'' synthesized
features are to the original features. The intuition here is that
closeness in feature space is indicative of an object-identity preserving 
synthesis.
In principle, we could simply evaluate $\|\phi_i^k(\mathbf{x})-\mathbf{x}\|^2$, 
however, the 2-norm is hard to interpret. Instead, we compute the 
Pearson correlation coefficient $\rho$ between each original feature and 
its synthesized variants, \ie, 
$\rho(\mathbf{x},\phi_i^k(\mathbf{x}))$. As $\rho$ ranges from $[-1,1]$, 
high values indicate a strong linear relationship to the original features.
Results are reported in Table~\ref{table:ednnonseen}. Similar to our 
previous results for MAE, we observe that $\rho$, when averaged over 
all objects, is slightly lower for objects that did not appear in the 
training data. This decrease in correlation, however, is relatively 
small. 
 
\begin{table}[t!]
\scriptsize
\centering{
\begin{tabular}{cr|cccc}
\hline
& \textbf{Object} & $\rho$ & \textbf{D} (MAE [m])  & $\rho$ & \textbf{P} (MAE [deg])    \\ \hline
\multirow{19}{*}{\begin{sideways}\textit{Seen} objects, see Table~\ref{table:maeCOR} \end{sideways}} 
&\texttt{bathtub} 	& 0.75 & 0.10 & 0.68 & 3.99 \\ 
&\texttt{bed} 		& 0.81 & 0.07 & 0.82 & 3.30 \\ 
&\texttt{bookshelf} & 0.80 & 0.06 & 0.79 & 3.36 \\ 
&\texttt{box} 		& 0.74 & 0.08 & 0.74 & 4.44 \\ 
&\texttt{chair} 	& 0.73 & 0.07 & 0.71 & 3.93 \\ 
&\texttt{counter} 	& 0.76 & 0.08 & 0.77 & 3.90 \\
&\texttt{desk} 		& 0.75 & 0.07 & 0.74 & 3.93 \\ 
&\texttt{door} 		& 0.67 & 0.10 & 0.63 & 4.71 \\ 
&\texttt{dresser} 	& 0.79 & 0.08 & 0.77 & 4.12 \\
&\texttt{garbage bin}& 0.76 & 0.07 & 0.76 & 5.30 \\ 
&\texttt{lamp} 		 & 0.82 & 0.08 & 0.79 & 4.83 \\
&\texttt{monitor} 	 & 0.82 & 0.06 & 0.80 & 3.34 \\ 
&\texttt{night stand}& 0.80 & 0.07 & 0.78 & 4.00 \\
&\texttt{pillow} 	& 0.80 & 0.08 & 0.81 & 3.87 \\ 
&\texttt{sink}		& 0.75 & 0.11 & 0.76 & 4.00 \\ 
&\texttt{sofa} 		& 0.78 & 0.08 & 0.78 & 4.29 \\ 
&\texttt{table} 	& 0.75 & 0.07 & 0.74 & 4.10 \\
&\texttt{tv} 		& 0.78 & 0.08 & 0.72 & 4.66 \\
&\texttt{toilet} 	& 0.80 & 0.10 & 0.81 & 3.70 \\ \hline
& $\varnothing$ 
& \cellcolor{black!10}{\textbf{0.77}} 
& \cellcolor{black!10}{\textbf{0.08}} 
& \cellcolor{black!10}{\textbf{0.76}} 
& \cellcolor{black!10}{\textbf{4.10}} \\
\hline
\multirow{10}{*}{ \begin{sideways}\textit{Unseen} objects ($\mathcal{T}_1$) \end{sideways} } 
&\texttt{picture} 	& 0.67 & 0.08 & 0.65 & 5.13 \\ 
&\texttt{ottoman} 	& 0.70 & 0.09 & 0.70 & 4.41 \\
&\texttt{whiteboard}& 0.67 & 0.12 & 0.65 & 4.43 \\ 
&\texttt{fridge} 	& 0.69 & 0.10 & 0.68 & 4.48 \\ 
&\texttt{counter} 	& 0.76 & 0.08 & 0.77 & 3.98 \\  
&\texttt{books} 	& 0.74 & 0.08 & 0.73 & 4.26 \\ 
&\texttt{stove} 	& 0.71 & 0.10 & 0.71 & 4.50 \\  
&\texttt{cabinet} 	& 0.74 & 0.09 & 0.72 & 3.99 \\ 
&\texttt{printer} 	& 0.73 & 0.08 & 0.72 & 4.59 \\ 
&\texttt{computer} 	& 0.81 & 0.06 & 0.80 & 3.73 \\ \hline
& $\varnothing$ 
& \cellcolor{black!10}{0.72} 
& \cellcolor{black!10}{0.09} 
& \cellcolor{black!10}{0.71} 
& \cellcolor{black!10}{4.35} \\
\hline
\end{tabular}}
\caption{\label{table:ednnonseen} Assessment of $\phi_i^k$ \wrt 
(1) Pearson correlation $(\rho)$ of \emph{synthesized} and 
\emph{original} features and (2) mean MAE of predicted
attribute values of synthesized features, $\gamma(\phi_i^k(\mathbf{x}))$,  
\wrt the desired attribute values $t$. \textbf{D} indicates
\texttt{Depth}-aug. features (MAE in [m]); \textbf{P} indicates \texttt{Pose}-aug.
features (MAE in [deg]).}
\vspace{-0.3cm}
\end{table}


In summary, we conclude that these results warrant the use of
$\phi_i^k$ on feature descriptors from object classes that have 
\emph{not} appeared in the training corpus. This enables us to 
test $\phi_i^k$ in transfer learning setups, as we will see in the 
following one-shot experiments of Secs.~\ref{subsection:one-shot}
and \ref{section:exp_scenes}.

\subsection{One-shot object recognition}
\label{subsection:one-shot}

First, we demonstrate the utility of our approach on
the task of one-shot object recognition in a transfer
learning setup. Specifically, we aim to learn attribute-guided 
augmenters $\phi_i^k$ from instances of object classes
that are available in an external, annotated database (in 
our case, SUN RGB-D). We denote this collection of object 
classes as our \emph{source classes} $\mathcal{S}$. 
Given one instance from a collection of completely different 
object classes, denoted as the \emph{target classes} 
$\mathcal{T}$, we aim to train a discriminant classifier 
$C$ on $\mathcal{T}$, \ie, $C: \mathcal{X} \rightarrow \{1,\ldots,|\mathcal{T}|\}$. 
In this setting,  $\mathcal{S} \cap \mathcal{T} =
\emptyset$. Note that no attribute annotations for instances 
of object classes in $\mathcal{T}$ are available. 
This can be considered a variant of 
transfer learning, since we transfer knowledge from 
object classes in $\mathcal{S}$ to instances of object
classes in $\mathcal{T}$, \emph{without} any prior knowledge 
about $\mathcal{T}$.

\vskip0.5ex
\noindent
\textbf{Setup.} We evaluate one-shot object recognition
performance on three collections of previously unseen 
object classes in the following setup: First, we randomly 
select two sets of 10 object classes and ensure
that each object class has at least 100 samples in the 
testing split of SUN RGB-D. We further ensure that no 
object class is in $\mathcal{S}$. This guarantees (1) that 
we have never seen the image, nor (2) the object class during 
training. Since, SUN RGB-D does not have object-centric images, we 
use the ground-truth bounding boxes to obtain the actual 
object crops. This allows us to tease out the benefit of 
augmentation without having to deal with confounding factors such 
as background noise. The two sets of object classes are denoted 
$\mathcal{T}_1$\footnote{\scriptsize $\mathcal{T}_1$ = \{\texttt{picture}, \texttt{whiteboard}, 
\texttt{fridge}, \texttt{counter}, \texttt{books}, \texttt{stove}, \texttt{cabinet}, \texttt{printer}, \texttt{computer}, \texttt{ottoman}\}}
and $\mathcal{T}_2$\footnote{\scriptsize$\mathcal{T}_2$ = 
\{\texttt{mug}, 
\texttt{telephone}, 
\texttt{bowl}, 
\texttt{bottle}, 
\texttt{scanner}, 
\texttt{microwave}, 
\texttt{coffee table}, 
\texttt{recycle bin}, 
\texttt{cart}, 
\texttt{bench}\}}. 
We additionally compile a third set of target classes 
$\mathcal{T}_3 = \mathcal{T}_1 \cup \mathcal{T}_2$ and
remark that $\mathcal{T}_1 \cap \mathcal{T}_2 = \emptyset$. 
Consequently, we have two 10-class problems and one 20-class 
problem. For each object image in $\mathcal{T}_i$, we
then collect RCNN \texttt{FC7} features. 

As a \emph{Baseline}, 
we ``train'' a linear C-SVM (on 1-norm normalized features) 
using only the single instances of each object class in 
$\mathcal{T}_i$ (SVM cost fixed to 10). Exactly the same parameter 
settings of the SVM are then used to train on the single 
instances + features synthesized by AGA. We repeat the 
selection of one-shot instances 500 times and report
the average recognition accuracy. For comparison, we 
additionally list 5-shot recognition results in 
the same setup.

\vskip0.5ex
\noindent
\textit{Remark.} The design of this experiment is  
similar to \cite[Section 4.3.]{Peng15a}, with the exceptions  
that we (1) \emph{do not} detect objects, (2) augmentation
is performed in feature space and (3) no 
object-specific information is available. The latter is 
important, since \cite{Peng15a} assumes the existence of 
3D CAD models for objects in $\mathcal{T}_i$ from which 
synthetic images can be rendered. In our case, augmentation
\emph{does not} require any a-priori information about the objects
classes.

\begin{table}[t!]
\footnotesize
\centering{
\begin{tabular}{rcccc}
\hline
& \textbf{Baseline} &  \texttt{AGA}+\textbf{D} & \texttt{AGA}+\textbf{P} &  \texttt{AGA}\textbf{+D}\textbf{+P} \\
\hline
& \multicolumn{4}{c}{\emph{One-shot}}\\
\hline
$\mathcal{T}_1$ (10)&  33.74	
					&  \cellcolor{green!30}{38.32~\checkmark} 
					&  \cellcolor{green!05}{37.25~\checkmark} 
					&  \cellcolor{green!60}{39.10~\checkmark}\\
$\mathcal{T}_2$ (10)&  23.76 	
					&  \cellcolor{green!30}{28.49~\checkmark}
					&  \cellcolor{green!05}{27.15~\checkmark}
					&  \cellcolor{green!60}{30.12~\checkmark}\\
$\mathcal{T}_3$ (20)&  22.84	
					&  \cellcolor{green!30}{25.52~\checkmark}
					&  \cellcolor{green!05}{24.34~\checkmark	}				
					&  \cellcolor{green!30}{26.67~\checkmark}\\ 					
					\hline
					& \multicolumn{4}{c}{\emph{Five-shot}}\\
					\hline
$\mathcal{T}_1$ (10)&  50.03
					&  \cellcolor{green!30}{55.04~\checkmark}
					&  \cellcolor{green!05}{53.83~\checkmark}
					&  \cellcolor{green!60}{56.92~\checkmark} \\
$\mathcal{T}_2$ (10)& 36.76   
					& \cellcolor{green!30}{44.57~\checkmark} 
					& \cellcolor{green!05}{42.68~\checkmark}
					& \cellcolor{green!60}{47.04~\checkmark}\\
$\mathcal{T}_3$ (20)& 37.37    
					& \cellcolor{green!30}{40.46~\checkmark} 
					& \cellcolor{green!05}{39.36~\checkmark}
					& \cellcolor{green!60}{42.87~\checkmark}\\
					\hline					
\end{tabular}}
\caption{\label{table:oneshot} \emph{Recognition accuracy} (over 500 trials) 
for three object recognition tasks; \emph{top}: one-shot, \emph{bottom}: five-shot.
Numbers in parentheses indicate the \#classes.
A '\checkmark' indicates that the result is statistically different (at 5\% sig.) 
from the \emph{Baseline}. +\textbf{D} indicates adding \texttt{Depth}-aug. 
features to the one-shot instances; +\textbf{P} indicates addition of \texttt{Pose}-aug. features
and +\textbf{D}, \textbf{P} denotes adding a combination of \texttt{Depth}-/\texttt{Pose}-aug. 
features.}
\end{table}

\vskip0.5ex
\noindent
\textbf{Results.} Table~\ref{table:oneshot} lists the classification accuracy for the different sets of one-shot training data. \emph{First}, using original one-shot instances augmented by \texttt{Depth}-guided features (+\textbf{D}); 
\emph{second}, using original features + \texttt{Pose}-guided features 
(+\textbf{P}) and \emph{third}, a combination of both (+\textbf{D}, \textbf{P});
In general, we observe that adding AGA-synthesized features improves 
recognition accuracy over the \emph{Baseline} in all cases. For \texttt{Depth}-augmented
features, gains range from 3-5 percentage points, for \texttt{Pose}-augmented 
features, gains range from 2-4 percentage points on average. We attribute
this effect to the difficulty in predicting object pose from 2D data, as can
be seen from Table~\ref{table:maeCOR}. Nevertheless, in both augmentation settings, 
the gains are statistically significant (\wrt the \emph{Baseline}), as evaluated by a Wilcoxn rank sum test
for equal medians \cite{Gibbons2011a} at $5\%$ significance (indicated by '\checkmark' 
in Table~\ref{table:oneshot}). Adding both \texttt{Depth}- \emph{and} 
\texttt{Pose}-augmented features to the original one-shot features achieves the greatest improvement in recognition accuracy, 
ranging from 4-6 percentage points. This indicates that
information from depth and pose is complementary and allows for better 
coverage of the feature space. Notably, we also experimented with the
metric-learning approach of Fink~\cite{Fink04a} which only led to negligible gains over the \emph{Baseline} (\eg, 33.85\% on $\mathcal{T}_1$).

\vskip0.5ex
\noindent
\textbf{Feature analysis/visualization.}
To assess the nature of feature synthesis, we backpropagate through RCNN layers the 
gradient \wrt the 2-norm between an original and a synthesized feature vector. The 
strength of the input gradient indicates \emph{how much each pixel of the object must 
change} to produce a proportional change in depth/pose of the sample.
\begin{figure}[t!]
\centering{
\includegraphics[width=\columnwidth]{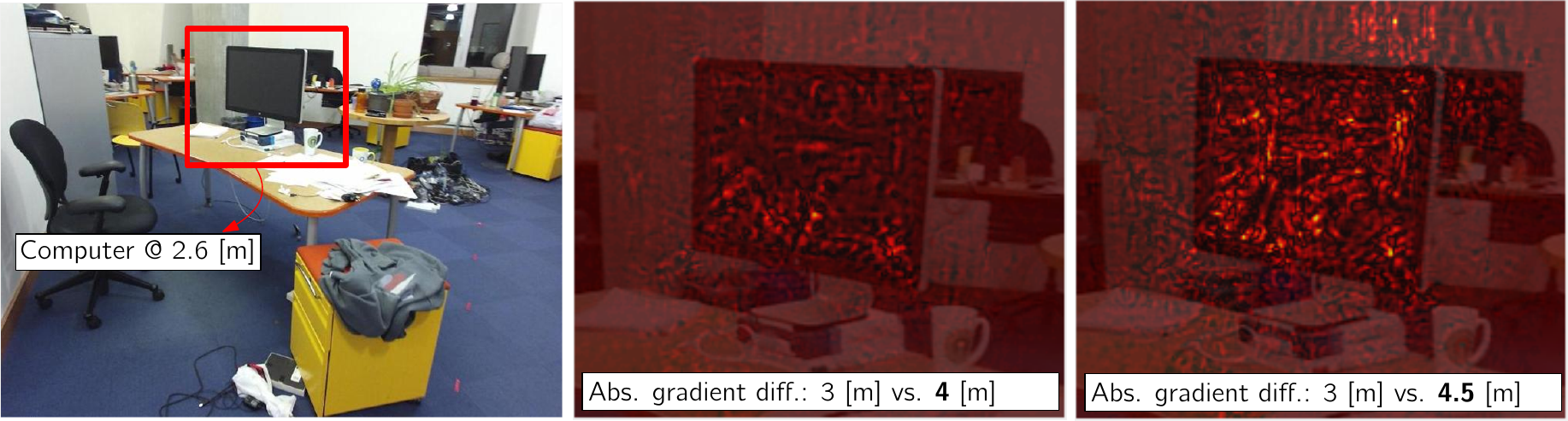}}
\caption{\label{fig:gradmag}Illustration of the 
difference in \emph{gradient magnitude} when backpropagating (through RCNN) 
the 2-norm of the difference between an original and a synthesized 
feature vector for an increasing desired change in depth, \ie, 
3[m] \vs 4[m] (\emph{middle}) and 3[m] \vs 4.5[m] (\emph{right}).}
\end{figure}
As can be seen in the example of Fig.~\ref{fig:gradmag}, a \emph{greater} desired 
change in depth invokes a \emph{stronger} gradient on the monitor.
\emph{Second}, we ran a \emph{retrieval experiment}: 
we sampled 1300 instances of 10 (unseen) object classes 
($\mathcal{T}_1$) and synthesized features for each 
instance \wrt depth. Synthesized 
features were then used for retrieval on the original 
1300 features. This allows to assess if synthesized 
features (1) allow to retrieve instances of the \emph{same class}
(Top-1 acc.) and (2) of the desired attribute value. 
The latter is measured by the coefficient of determination 
($R^2$). As seen in Table~\ref{table:retrieval}, the $R^2$ scores indicate that 
we can actually retrieve instances with the desired 
attribute values. Notably, even in cases
where $R^2 \approx 0$ (\ie, the linear model does
not explain the variability), the results still show decent 
Top-1 acc., revealing that synthesis \emph{does not alter 
class membership}.
\begin{table}[h!]
\centering
\begin{scriptsize}
\begin{tabular}{r|cc||r|cc}
\hline
\textbf{Object} & Top-1 & $R^2$ & \textbf{Object} & Top-1 & $R^2$\\
 \hline
\texttt{picture} 	& 0.33 & 0.36 	& \texttt{whiteboard} 	& 0.12 & 0.30 \\
\texttt{fridge} 	& 0.26 & 0.08 	& \texttt{counter} 		& 0.64 & 0.18 \\
\texttt{books} 	 	& 0.52 & 0.07 	& \texttt{stove} 		& 0.20 & 0.13 \\
\texttt{cabinet} 	& 0.57 & 0.27	& \texttt{printer} 		& 0.31 & 0.02 \\
\texttt{computer}	& 0.94 & 0.26    & \texttt{ottoman} 		& 0.60 & 0.12 \\
\hline
\end{tabular}
\end{scriptsize}
\caption{\label{table:retrieval}\emph{Retrieval results} for 
unseen objects ($\mathcal{T}_1$) when querying with synthesized features of 
varying depth. Larger
$R^2$ values indicate a stronger linear relationship ($R^2 \in [0,1]$) 
to the depth values of retrieved instances.}
\end{table}


\subsection{Object-based one-shot scene recognition}
\label{section:exp_scenes}

\begin{table}[t!]
\footnotesize
\centering{
\begin{tabular}{lc}
\hline
\textbf{Method} & \textbf{Accuracy}~[\%] \\
\hline
\texttt{max.\ pool} (\textit{Baseline}) & 13.97 \\
\texttt{AGA\ FV} (\textbf{+D}) & \cellcolor{green!10}{\textbf{15.13}}\\
\texttt{AGA\ FV} (\textbf{+P}) & \cellcolor{green!10}{\textbf{14.63}} \\
\texttt{AGA\ CL-1} (\textbf{+D}, max.) & \cellcolor{green!20}{\textbf{16.04}} \\
\texttt{AGA\ CL-2} (\textbf{+P}, max.) & \cellcolor{green!20}{\textbf{15.52}} \\
\texttt{AGA\ CL-3} (\textbf{+D}, \textbf{+P}, max.) & \cellcolor{green!30}{\textbf{16.32}} \\
\hline
\texttt{Sem-FV}~\cite{Dixit15a} & 32.75 \\
\texttt{AGA Sem-FV} & \cellcolor{green!50}{\textbf{34.36}}\\
\hline
\texttt{Places}~\cite{Zhou14a} & 51.28 \\
\texttt{AGA Places} & \cellcolor{green!50}{\textbf{52.11}}\\
\hline
\end{tabular}}
\caption{\label{table:oneshot_scenes}
\textit{One-shot classification} on 25 indoor scene classes
\cite{dset:MITIndoor}: \{auditorium, bakery, bedroom, bookstore, children room, classroom, computer room, concert hall, corridor, dental office, dining room, hospital room, laboratory, library, living room, lobby, meeting room, movie theater, nursery, office, operating room, pantry, restaurant\}. For \texttt{Sem-FV}~\cite{Dixit15a}, we use ImageNet CNN features extracted at one image scale.}
\vspace{-0.3cm}
\end{table}

\noindent
\textbf{Motivation.} We can also use AGA for a different type of 
transfer, namely the transfer from object detection networks to 
one-shot scene recognition. Although, object detection is a challenging 
task in itself, significant progress is made, every year, 
in competitions such as the ImageNet challenge. Extending the gains 
in object detection to other related problems, such as scene recognition, 
is therefore quite appealing. A system that uses 
an accurate object detector such as an RCNN~\cite{Girshick15a} to perform 
scene recognition, could generate comprehensive annotations for an 
image in one forward pass. An object detector that supports one-shot scene 
recognition could do so with the least amount of additional data. 
It must be noted that such systems are different from object recognition 
based methods such as~\cite{Gong14a,Dixit15a,Cimpoi15a}, where explicit 
detection of objects is not necessary. They apply filters from object 
recognition CNNs to several regions of images and extract features from 
all of them, whether or not an object is found. The data available to them is therefore enough to learn complex descriptors such as Fisher vectors (FVs). A detector, on the other hand, may produce very few features from an image, based on the number of objects found. 
AGA is tailor-made for such scenarios where 
features from an RCNN-detected object can be augmented.

\vskip0.5ex
\noindent
\textbf{Setup.}
To evaluate AGA in this setting, we select a 25-class subset of 
MIT Indoor~\cite{dset:MITIndoor}, which may contain 
objects that the RCNN is trained for. The reason for this choice 
is our reliance on a detection CNN, which has a vocabulary of 
19 objects from SUN RGB-D. At present, this is the largest such 
dataset that provides objects and their 3D attributes. The system 
can be extended easily to accommodate more scene classes if a 
larger RGB-D object dataset becomes available. As the RCNN produces 
very few detections per scene image, the best approach, without 
augmentation, is to perform pooling of RCNN features from proposals 
into a fixed-size representation. We used max-pooling as our 
\textit{baseline}. Upon augmentation, using predicted depth/ pose, an image 
has enough RCNN features to compute a GMM-based FV. 
For this, we use the experimental settings 
in~\cite{Dixit15a}. The FVs are denoted as 
\texttt{AGA~FV(\textbf{+D})} and \texttt{AGA~FV(\textbf{+P})}, based on 
the attribute used to guide the augmentation. 
As classifier, we use a linear C-SVM with fixed parameter (C). 

\vskip0.5ex
\noindent
\textbf{Results.} Table~\ref{table:oneshot_scenes} lists the 
avgerage one-shot recognition accuracy over multiple iterations. The benefits of AGA are 
clear, as both aug. FVs perform better than the max-pooling baseline by 0.5-1\% points. Training on a combination (concatenated vector) of the augmented FVs and max-pooling, denoted as \texttt{AGA\ CL-1}, \texttt{AGA\ CL-2} and \texttt{AGA\ CL-3} further improves by about 1-2\% points. Finally, we combined our aug.~FVs with the 
state-of-the-art semantic FV of~\cite{Dixit15a} and Places CNN features~\cite{Zhou14a} for one-shot classification. 
Both combinations, denoted \texttt{AGA\ Sem-FV} and \texttt{AGA\ Places}, improved by a non-trivial margin ($\sim$1\% points).  

\section{Discussion}
\label{section:discussion}

We presented an approach toward attribute-guided 
augmentation in feature space. Experiments
show that object attributes, such as pose / depth, are
beneficial in the context of one-shot recognition, \ie,
an extreme case of limited training data. Notably,
even in case of mediocre performance of the attribute
regressor (\eg, on pose), results indicate that 
synthesized features can still supply useful 
information to the classification process.
While we do use bounding boxes to extract object crops from 
SUN RGB-D in our object-recognition experiments, this is only 
done to clearly tease out the effect of augmentation. 
In principle, as our encoder-decoder is trained
in an \emph{object-agnostic} manner, no external knowledge 
about classes is required. 

As SUN RGB-D exhibits high variability in the range 
of both attributes, augmentation along these dimensions can 
indeed help classifier training. However, when variability
is limited, \eg, under controlled acquisition 
settings, the gains may be less apparent. In that case,
augmentation with respect to other object attributes 
might be required.

Two aspects are specifically interesting for future work. \emph{First},
replacing the attribute regressor for pose with a specifically
tailored component will potentially improve learning of the
synthesis function(s) $\phi_i^k$ and lead to
more realistic synthetic samples. \emph{Second}, we conjecture
that, as additional data with more annotated object classes 
and attributes becomes available (\eg, \cite{Borji16a}), 
the encoder-decoder can leverage more diverse samples and 
thus model feature changes with respect to the 
attribute values more accurately. 

\vskip1ex
\noindent
\textbf{Acknowledgments.}
This work is supported by NSF awards IIS-1208522, CCF-0830535, ECCS-1148870 and a 
generous donation of GPUs from Nvidia.

\newpage
{\small
\bibliographystyle{ieee}
\bibliography{egbib,halucination}

\begin{thebibliography}{10}\itemsep=-1pt

\bibitem{Bengio09a}
Y.~Bengio.
\newblock Learning deep architectures for {AI}.
\newblock {\em Found. Trends Mach. Learn.}, 2(1):1--127, 2009.

\bibitem{Borji16a}
A.~Borji, S.~Izadi, and L.~Itti.
\newblock {iLab-20M}: A large-scale controlled object dataset to investigate
  deep learning.
\newblock In {\em CVPR}, 2016.

\bibitem{Rasmussen05a}
C.~W. C.E.~Rasmussen.
\newblock {\em Gaussian Processes for Machine Learning}.
\newblock The MIT Press, 2005.

\bibitem{Charalambous16a}
C.~Charalambous and A.~Bharath.
\newblock A data augmentation methodology for training machine/deep learning
  gait recognition algorithms.
\newblock In {\em BMVC}, 2016.

\bibitem{Chatfield14}
K.~Chatfield, K.~Simonyan, A.~Vedaldi, and A.~Zisserman.
\newblock Return of the devil in the details: Delving deep into convolutional
  nets.
\newblock In {\em BMVC}, 2014.

\bibitem{Cimpoi15a}
M.~Cimpoi, S.~Maji, and A.~Vedaldi.
\newblock Deep filter banks for texture recognition and segmentation.
\newblock In {\em CVPR}, 2015.

\bibitem{Clevert16a}
D.-A. Clevert, T.~Unterhiner, and S.~Hochreiter.
\newblock Fast and accurate deep network learning by exponential linear units
  ({ELU}s).
\newblock In {\em ICLR}, 2016.

\bibitem{Dixit15a}
M.~Dixit, S.~Chen, D.~Gao, N.~Rasiwasia, and N.~Vasconcelos.
\newblock Scene classification with semantic {Fisher} vectors.
\newblock In {\em CVPR}, 2015.

\bibitem{Donahue14a}
J.~Donahue, Y.~Jia, O.~Vinyals, J.~Huffman, N.~Zhang, E.~Tzeng, and T.~Darrell.
\newblock {DeCAF}: A deep convolutional activation feature for generic visual
  recognition.
\newblock In {\em ICML}, 2014.

\bibitem{Drucker97a}
H.~Drucker, C.~Burges, L.~Kaufman, and A.~Smola.
\newblock Support vector regression machines.
\newblock In {\em NIPS}, 1997.

\bibitem{Fan08a}
R.-E. Fan, K.-W. Chang, C.-J. Hsieh, X.-R. Wang, and C.~J. Lin.
\newblock {LIBLINEAR}: A library for large linear classification.
\newblock {\em JMLR}, 9(8):1871--1874, 2008.

\bibitem{Fink04a}
M.~Fink.
\newblock Object classification from a single example utilizing relevance
  metrics.
\newblock In {\em NIPS}, 2004.

\bibitem{Gibbons2011a}
J.~Gibbons and S.~Chakraborti.
\newblock {\em Nonparametric Statistical Inference}.
\newblock Chapman \& Hall/CRC Press, 5th edition, 2011.

\bibitem{Girshick15a}
R.~Girshick.
\newblock Fast {R-CNN}.
\newblock In {\em ICCV}, 2015.

\bibitem{Gong14a}
Y.~Gong, L.~Wang, R.~Guo, and S.~Lazebnik.
\newblock Multi-scale orderless pooling of deep convolutional activation
  features.
\newblock In {\em ECCV}, 2014.

\bibitem{Hauberg16a}
S.~Hauberg, O.~Freifeld, A.~Boensen, L.~Larsen, J.~F. III, and L.~Hansen.
\newblock Dreaming more data: Class-dependent distributions over
  diffeomorphisms for learned data augmentation.
\newblock In {\em AISTATS}, 2016.

\bibitem{Ioffe15a}
S.~Ioffe and C.~Szegedy.
\newblock Batch normalization: Accelerating deep network training by reducing
  internal covariate shift.
\newblock In {\em ICML}, 2015.

\bibitem{Jaderberg15a}
M.~Jaderberg, K.~Simonyan, A.~Zisserman, and K.~Kavukcuoglu.
\newblock Spatial transformer networks.
\newblock In {\em NIPS}, 2015.

\bibitem{Kingma15a}
D.~Kingma and J.~Ba.
\newblock Adam: A method for stochastic optimization.
\newblock In {\em ICLR}, 2015.

\bibitem{Krizhevsky12a}
A.~Krizhevsky, I.~Sutskever, and G.~E. Hinton.
\newblock {Imagenet} classification with deep convolutional neural networks.
\newblock In {\em NIPS}, 2012.

\bibitem{Kwitt16a}
R.~Kwitt, S.~Hegenbart, and M.~Niethammer.
\newblock One-shot learning of scene locations via feature trajectory transfer.
\newblock In {\em CVPR}, 2016.

\bibitem{Liu15a}
F.~Liu, C.~Shen, and G.~Lin.
\newblock Deep convolutional neural fields for depth estimation from a single
  image.
\newblock In {\em CVPR}, 2015.

\bibitem{Miller00a}
E.~Miller, N.~Matsakis, and P.~Viola.
\newblock Learning from one-example through shared density transforms.
\newblock In {\em CVPR}, 2000.

\bibitem{Nair10a}
V.~Nair and G.~Hinton.
\newblock Rectified linear units improve restricted boltzmann machines.
\newblock In {\em ICML}, 2010.

\bibitem{Peng15a}
X.~Peng, B.~Sun, K.~Ali, and K.~Saenko.
\newblock Learning deep object detectors from 3d models.
\newblock In {\em ICCV}, 2015.

\bibitem{dset:MITIndoor}
A.~Quattoni and A.~Torralba.
\newblock Recognizing indoor scenes.
\newblock In {\em CVPR}, 2009.

\bibitem{Ren15a}
S.~Ren, K.~He, R.~Girshick, and J.~Sun.
\newblock Faster {R-CNN}: Towards real-time object detection.
\newblock In {\em NIPS}, 2015.

\bibitem{Rogez16a}
G.~Rogez and C.~Schmid.
\newblock {MoCap}-guided data augmentation for {3D} pose estimation in the
  wild.
\newblock {\em CoRR}, abs/1607.02046, 2016.

\bibitem{Fergus14a}
P.~Sermanet, D.~Eigen, X.~Zhang, M.~Mathieu, R.~Fergus, and Y.~LeCun.
\newblock {OverFeat}: Integrated recognition, localization and detection using
  convolutional networks.
\newblock In {\em ICLR}, 2014.

\bibitem{Simonyan14a}
K.~Simonyan and A.~Zisserman.
\newblock Very deep convolutional networks for large-scale image recognition.
\newblock {\em CoRR}, abs/1409.1556, 2014.

\bibitem{Song15a}
S.~Song, S.~Lichtenberg, and J.~Xiao.
\newblock {SUN RGB-D}: A {RGB-D} scene understanding benchmark suite.
\newblock In {\em CVPR}, 2015.

\bibitem{Srivastava14a}
N.~Srivastava, G.~Hinton, A.~Krizhevsky, I.~Sutskever, and R.~Salakhutdinov.
\newblock Dropout: A simple way to prevent neural networks from overfitting.
\newblock {\em JMLR}, 15:1929−1958, 2014.

\bibitem{Su15a}
H.~Su, C.~Qi, Y.~Li, and L.~Guibas.
\newblock Render for {CNN}: Viewpoint estimation in images using cnns trained
  with rendered 3d model views.
\newblock In {\em ICCV}, 2015.

\bibitem{Szegedy15a}
C.~Szegedy, W.~Liu, Y.~Jia, P.~Sermanet, S.~Reed, D.~Anguelov, D.~Erhan,
  V.~Vanhoucke, and A.~Rabinovich.
\newblock Going deeper with convolutions.
\newblock In {\em CVPR}, 2015.

\bibitem{Torralba11a}
A.~Torralba and A.~Efros.
\newblock Unbiased look at dataset bias.
\newblock In {\em CVPR}, 2011.

\bibitem{Uijlings13a}
J.~Uijlings, K.~van~de Sande, T.~Gevers, and A.~Smeulders.
\newblock Selective search for object recognition.
\newblock {\em IJCV}, 104(2):154--171, 2013.

\bibitem{Zeiler14a}
M.~Zeiler and R.~Fergus.
\newblock Visualizing and understanding convolutional networks.
\newblock In {\em ECCV}, 2014.

\bibitem{Zhou14a}
B.~Zhou, A.~Lapedriza, J.~Xiao, A.~Torralba, and A.~Oliva.
\newblock Learning deep features for scene recognition using {Places} database.
\newblock In {\em NIPS}, 2014.

\end{thebibliography}
}

\end{document}